# Enhancing Computational Cognitive Architectures with LLMs: A Case Study

Ron Sun

## Abstract

Computational cognitive architectures are broadly scoped models of the human mind that combine different psychological functionalities (as well as often different computational methods for these different functionalities) into one unified framework. They structure them in a psychologically plausible and validated way. However, such models thus far have only limited computational capabilities, mostly limited by the computational tools and techniques that were adopted. More recently, LLMs have proved to be more capable computationally than any other tools. Thus, in order to deal with both real-world complexity and psychological realism at the same time, incorporating LLMs into cognitive architectures naturally becomes an important task. In the present article, a synergistic combination of the Clarion cognitive architecture and LLMs is discussed as a case study. The implicit-explicit dichotomy that is fundamental to Clarion is leveraged for a seamless integration of Clarion and LLMs. As a result, computational power of LLMs is combined with psychological nicety of Clarion.

## 1. Introduction

A computational cognitive architecture, generally speaking, is a general-purpose computational theory of the human mind, resulting from empirical and theoretical work in cognitive science (especially computational psychology). It is a broadly scoped psychological model that combines different psychological functionalities (action, reasoning, memory, learning, motivation, metacognition, etc.) and often different computational methods for these different functionalities (symbolic, neural, etc.) into one unified framework. It structures these functionalities and methods in a psychologically plausible/realistic and reasonably well validated way. Given its computational nature, it is mechanistic and process based. Thus it can be detailed yet precise, shedding light on the human mental processes beyond what other approaches (e.g., verbal-conceptual theories, or pure mathematical theories) can. Note that here we are concerned only with psychologically realistic computational cognitive architectures that are meant to be rigorous models of the human mind and have been validated empirically (e.g., Taatgen & Anderson, 2023).

One example is Clarion, which has been developed since the 1990s (see, e.g., Sun, 2002, 2016). It has been extensively justified by, and validated against, psychological data, findings, and theoretical constructs, and accounts for a wide range of psychological data and findings (more so than other existing cognitive architectures). One important theoretical underpinning for Clarion is the dual-process theories from the psychology and philosophy literatures, which led to the overall two-level structuring of Clarion in a hybrid neuro-symbolic way (as will be elaborated later).

In general, the field of cognitive science has been developing computational models since the 1970s, which are meant to capture human behavior in a wide range of activities, but their actual capabilities have not been keeping up with the advances in new technology. For instance, some cognitive architectures, initially conceived in the 1970s, relied on technology available at that time. Despite the fact that some new techniques have been incorporated, their overall frameworks, as well as some of their details, often seem out of date. Therefore, it may be argued that cognitive models need an overhaul to incorporate new advances in AI and in computer science.

For instance, shortcomings of the current version of Clarion include, for example, its limited knowledge, especially its limited implicit knowledge. Implicit knowledge should include a vast store of intuition and instinct, beyond what a simple three-layer Backpropagation network can hold (more on this later). Another shortcoming is the limited interaction between implicit and explicit processes: Although Clarion developed some essential algorithms for that, more needs to be done. Equally important, Clarion currently lacks the capability for natural language communication. Therefore, it needs to be updated with new computing techniques.

On the other hand, more recently, LLMs (and Transformer-based models in general) have proved to be more computationally capable than any other computational models or methods that have been seen thus far. Therefore, incorporating LLMs (or Transformer-based models) into cognitive architectures naturally becomes an important or even crucial task facing cognitive architecture research. This is because such incorporation is the best way forward for cognitive architectures to deal with both real-world complexity and psychological realism at the same time.

In the present article, as a case study, a synergistic combination of the Clarion cognitive architecture and LLMs is discussed. In particular, the implicit-explicit dichotomy fundamental to Clarion is leveraged for a natural and seamless integration of Clarion and LLMs. As a result, computational power of LLMs is combined with psychological nuances such as implicit-explicit interaction, various memory stores, intrinsic motivation, metacognitive regulation, and other useful facets of Clarion.

In the remainder of this article, first, some background on cognitive architectures and Clarion is presented. Then, various possibilities of integrating a cognitive architecture with LLMs are discussed, which leads to a new version of Clarion. The new Clarion is illustrated with examples and other descriptions. Finally, some general discussions conclude the paper.

## 2. Background on Cognitive Architectures

### 2.1. Why Cognitive Architectures?

Computational psychology is about understanding the human mind through developing computational models in a psychologically rigorous way. These models embody descriptions of psychological functioning in computer algorithms and programs. That is, they impute computational processes onto psychological functions. They thus lead to runnable computer programs. Detailed computational simulations and other operations can be carried out based on these models. Given the complexity of the human mind and the consequent behavioral complexity, precise, mechanistic, process-oriented computational models are necessary to explicate its intricate details. These also support practical applications of cognitive-psychological theories (Pew & Mavor, 1998).

Within this field, a *computational cognitive architecture* is an empirically (psychologically) grounded, comprehensive, computational framework for understanding the human mind (e.g., Taatgen & Anderson, 2023). The advantage is that it provides an essential framework that facilitates more detailed exploration of the mind. The many assumptions that it embodies may be based on empirical data, philosophical analysis, or computationally inspired hypotheses. Through embodying these constraints, a cognitive architecture narrows down possibilities and provides scaffolding structures. As mentioned earlier, Clarion is one such cognitive architecture (Sun, 2016).

## 2.2. Dual-process Theories and Neuro-symbolic Models

One foundational idea in Clarion is the implicit-explicit distinction. The distinction between implicit processes (also termed System 1) and explicit processes (also termed System 2) has been argued by many since the 1980s (e.g., Evans & Frankish, 2009; Macchi et al., 2016; Reber, 1989; Sun, 1994) and later popularized by Kahneman (2011). Various dual-process theories have been steadily gaining attention, in psychology, philosophy, and other fields, but that distinction in Clarion goes back right to the beginning (as in Sun, 1994).

Generally speaking, explicit processes are relatively accessible to consciousness, while implicit processes are less so. Explicit processing may be described as symbolic and rule-based to some significant extent, while implicit processing is more "associative" and "holistic". Empirical evidence and theoretical analysis in support of these points can be found in the literature (as cited above). Work on computational cognitive architectures has also demonstrated computationally the importance of dual processes (e.g., Helie & Sun, 2010; Sun et al., 2005).

In this regard, it has been argued that hybrid neuro-symbolic models may be necessary to address the dual processes of the human mind (see Sun, 1994, 2023). The neural (subsymbolic) side of a hybrid system may address low-level sensory-motor processes and other implicit processes, while the symbolic side can perform higher-level, more deliberative reasoning, planning, and reflection. These two sides may be distinct structurally and mechanistically but can work together in a synergistic way. It has been emphasized that such models should be structured rigorously in a psychologically motivated and justified way, based on the structures, mechanisms, and processes of human mental processes (e.g., Sun, 2023).

## 2.3. The Clarion Cognitive Architecture

### 2.3.1. Overall Structure

Clarion is meant to be a broad, mechanistic, process-based, psychological theory (Sun, 2002, 2016). Clarion consists of four major subsystems: the action-centered subsystem (ACS) for dealing with actions, involving procedural (i.e., action-oriented) knowledge; the non-action-centered subsystem (NACS) for memory and reasoning, involving declarative (i.e., factual) knowledge; the motivational subsystem (MS) for providing motivation for the other subsystems; and the metacognitive subsystem (MCS) for regulating other subsystems. Such an overall structure has been argued on the basis of empirical findings and theoretical analysis (Sun, 2002, 2016).

Each of these subsystems consists of two "levels". The top level carries out explicit processes; the bottom level carries out implicit processes (Sun, 1994, 2002). Representationally, one is symbolic and the other neural (e.g., Backpropagation networks). The two types of representations are inter-connected: The symbolic representations at the top level are in the forms of "chunk" nodes (each representing a concept, defined by a set of microfeatures, which, however, are located at the bottom level) and rules connecting these chunk nodes. Each chunk node is linked to the corresponding microfeature nodes at the bottom level; thus the two kinds of processes interact to generate combined outcomes. See Figure 1.

The flow of information among these subsystems may be roughly as follows (Sun et al., 20022): First, within the MS, situational inputs trigger internal motives, termed *drives*, on the basis of internal drive propensities. Then, a goal (an intention for action) within the MS is selected (by the MCS) on the basis of activated drives (for the sake of satisfying the drives). Action/meta-action selection occurs, within the ACS/MCS, on the basis of the goal selected and the situational inputs (to achieve the satisfaction of the drives). Reinforcement learning, which is fundamental to Clarion, then occurs in the ACS/MCS on the basis of the overall satisfaction of the drives (as reward), along with other forms of learning. Actions selected by the ACS/MCS may include directives to the NACS to perform certain reasoning (thus the NACS may be under the control of the ACS), in addition to actions that affect the external world.

Clarion has been well validated against psychological data, findings, and theories (see, e.g., Sun, 2016 for details). It is suitable as a framework for incorporating LLMs (as has already been explored, e.g., by Romero et al., 2023; Sun, 2024).

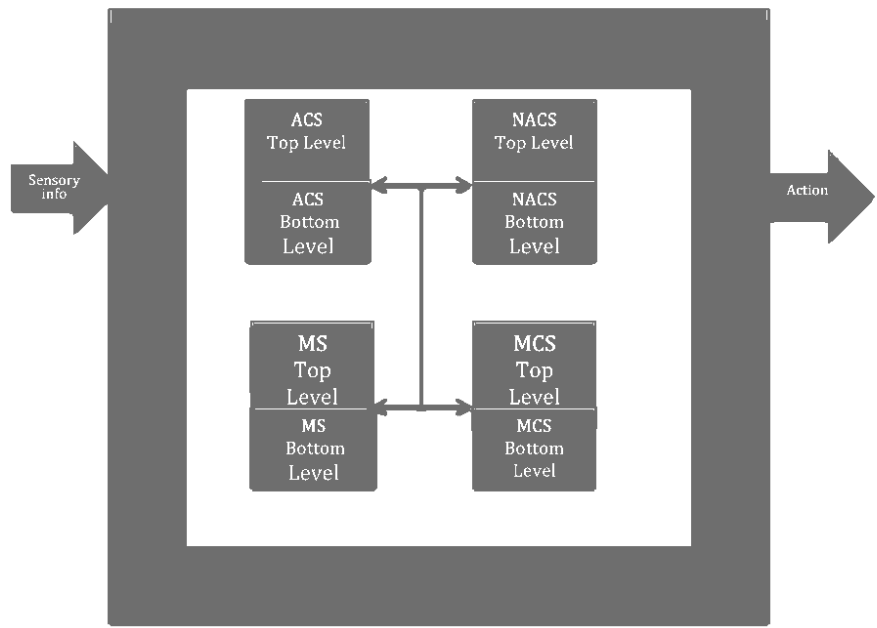

*Figure 1. The original Clarion cognitive architecture with its four major subsystems. See the text for explanations.*

### 2.3.2. Some Important Interactions

In this original Clarion, explicit processes are captured using symbolic representations (with chunk nodes representing concepts and rules connecting them). Implicit processes are captured using three-layer Backpropagation neural networks (involving microfeature nodes). (Limitations of such implementations will be addressed later.)

In this Clarion, implicit-explicit interaction has been explored computationally, both in the ACS and in the NACS (Sun, 2002, 2016), which has also been validated (through modeling and simulation of psychological data; see, e.g., Sun et al., 2001, 2005; Helie et al., 2010). One essential form of implicit-explicit interaction is what has been termed *top-down activation* — initiation of implicit processes at the bottom level by explicit processes at the top level, achieved through chunk nodes at the top level activating their corresponding microfeature nodes at the bottom level. The other form of interaction is *bottom-up activation* — activation of symbolic representations at the top level by implicit processes at the bottom level, through microfeatures at the bottom level activating their corresponding chunk nodes at the top level. In this way, the two levels affect each other. Implicit processes might be guided, in some way, by explicit processes

(e.g., by explicit instructions); alternatively, results of implicit processes might potentially be verified, refined, or explained by explicit processes in some way. These forms of interaction were well justified based on the literature.

Note that a simple way of integrating the outcomes across levels is combining the outputs of the two levels through stochastic selection, whereby the selection probabilities are determined based on relative performance of levels (Sun, 2016).

In relation to learning, one form of implicit-explicit interaction in Clarion is *top-down learning* (learning explicit knowledge first and then assimilating it into implicit processes), and another is *bottom-up learning* (learning implicit knowledge first and then learning explicit knowledge on that basis). See Sun (2016) for technical details of these forms of learning. In particular, bottom-up learning is based on statistical measures (such as information gain), while top-down learning is based on reinforcement-based fine-tuning where the top level provides "reinforcement". Along with other forms of learning (especially regular reinforcement learning, which is fundamental to Clarion as mentioned earlier), bottom-up and top-down learning enable incremental, continuous, real-time learning in Clarion from situations encountered.

In Clarion, there are a variety of memory across subsystems, often in both implicit and explicit forms. Among them, procedural memory (concerning procedural knowledge) is located within the ACS, while semantic and episodic memory are in the NACS (for retrieval of semantic and episodic information and for reasoning). In addition, working memory serves as (additional) inputs to both the ACS and the NACS. These types of memory are important for human cognition, as demonstrated in empirical research (e.g., Baddeley, 1986; Schacter, 1987; Tulving, 1983). They are also important to Clarion: They interact to accomplish psychological functionalities (Sun, 2016).

Yet another crucial form of interaction occurs among subsystems. For instance, the NACS is usually under the control of the ACS, so reasoning usually occurs in the service of action. For another example, the MS, with its motivational states, provides impetus for the ACS (and, by extension, the NACS), while the MCS regulates other subsystems based on the motivational states within the MS (for details, see Sun, 2016).

The capabilities of Clarion for capturing human psychological processes have been extensively demonstrated (see, e.g., Sun et al., 2001, 2005; Helie & Sun, 2010). However, it does have several serious shortcomings (which, incidentally, are also shared by other cognitive architectures; Taatgen & Anderson, 2023).

### 2.4. Current Shortcomings

Shortcomings of the current Clarion include (as touched upon earlier):
- A limited amount of knowledge, especially implicit knowledge
- Lack of natural language communication
- Limited interaction between implicit and explicit processes; in particular, the limited extent to which one's "inner dialogue" has been captured.

In terms of implicit knowledge, Reber (1989) pointed out that "implicitly acquired epistemic contents of mind are always richer and more sophisticated than what can be explicated" (p. 229). He pointed to the depth and the breadth of implicit knowledge. Dreyfus and Dreyfus (1987) emphasized the difference between "knowing-that" (explicit knowledge) and "knowing-how" (intuition): One is conscious (explicit), symbolic, and often slow and analytical, while the other is unconscious (implicit) and reactive but often instantaneous. Dreyfus and Dreyfus argued that, when one's intuition has developed to the point that one simply reacts (without resorting to explicit rules), knowledge is not stored symbolically but affects what one notices, what one expects, and what possibilities one ignores or focuses on. Such human know-how is powerful and vast in scope (Goldberg, 1983).

Correspondingly, we may examine Clarion in this light. Simple Backpropagation networks previously used in the bottom level of Clarion are internally inaccessible and fast, consistent with phenomenological analysis of human intuition and instinct. But they are, however, not capable of holding simultaneously a vast amount of implicit knowledge like humans do. Their capacities are limited by the nature of such simple models (with problems such as catastrophic forgetting, diminishing gradients, and so on). That is why better methods (such as Transformers and LLMs) are needed in Clarion. They are computationally necessary in order to deal with the complexity of the real world. Cognitive architectures, including Clarion, need to go beyond simulating laboratory tasks and into dealing with real-world complexities.

In terms of natural language communication (e.g., with external entities such as humans or human-like agents), such a capability is lacking in Clarion, the same in almost all currently existing computational cognitive architectures. Before the emergence of LLMs, natural language communication seemed a huge undertaking that would add a tremendous amount of seemingly irrelevant work to cognitive architecture research. With LLMs, the situation has changed: It seems that the capability of natural language processing (or at least an approximation of it) can indeed be part of cognitive architectures.

In terms of the "inner dialogue" mentioned earlier, Vygotsky (1962) provided some seminal ideas. In his view, one's internal thoughts develop when one internalizes outward utterances, so that inter-personal conversation is turned into intra-personal inner dialogue. However, he did not identify the internal interlocutors. According to Clarion, one such internal "interlocutor" may be one's own implicit mind — the part of the mind that is not directly accessible (in terms of its inner working). It is a black box that may generate thoughts, apparently without identifiable intermediate steps (Dreyfus & Dreyfus, 1987; Goldberg, 1983; Helie & Sun, 2010). Thus, one's explicit mind may need to engage it in a "dialogue". Helie and Sun (2010) elucidated such a dialogue in the case of creative problem solving: First, explicit processes prepare for problem solving by gathering and explicitly analyzing information and communicating it to the implicit mind; then, implicit processes work on the problem based on the information provided; after a solution idea emerges within implicit processes, explicit processes are informed and they in turn explicitly analyze and validate the idea; if the idea is inadequate or incorrect, the information is sent back to implicit processes and the above steps are repeated. However, such inner dialogue in other domains under many different circumstances needs to be elucidated and hopefully incorporated into cognitive architectures.

Besides such inner dialogue, many other kinds of implicit-explicit interaction may also be considered. For example, the idea of "veto", as highlighted by the work of Libet (1985), was that explicit mind was responsible for vetoing implicitly initiated actions. Relatedly, the idea of "default-intervention" as proposed by Evans (Evans & Frankish, 2009) was that implicit processes formed default responses, unless there was active intervention from explicit processes. Curran and Keele (1993) and Stadler (1995) showed that implicit processes could be influenced or controlled by explicit processes. In terms of learning, there are many variations of the interaction of the two types of processes, in addition to the top-down and bottom-up learning algorithms in Clarion discussed earlier. Although Clarion developed some essential algorithms for the interaction (Sun et al., 2001, 2005), more are needed to capture the full extent of the interaction. All of these issues should be considered in a new cognitive architecture.

# 3. Enhancing Clarion with LLMs

## 3.1. Complementarity

LLMs (and Transformers on which they are based) may have strengths complementary to Clarion: What is lacking in Clarion may be what LLMs are good at (Chang & Bergen, 2023; Sartori & Orrù, 2023; Yildirim & Paul, 2024; and so on).

First, natural language communication is the most apparent advantage of LLMs. It has been demonstrated that LLMs are capable of interpreting questions or requests in natural language and then generating responses in perfectly grammatical sentences. Therefore, it is expedient to rely on LLMs to handle this aspect for cognitive architectures (even though LLMs may only approximate human linguistic capabilities).

Second, beyond natural language, LLMs hold internally an enormous amount of knowledge (gleaned from texts of many different sources). LLMs, from closed-sourced ChatGPT to open-source alternatives such as Llama, have demonstrated their capacity of storing vast knowledge in an implicit form (i.e., embedded in a large number of weights). Not directly accessible, such knowledge can nevertheless be observed through interactions with them (e.g., through questions and answers). When properly utilized, LLMs can enhance almost any cognitive architectures in terms of quantity of knowledge.

Third, inputs and outputs from LLMs can be in various forms, beyond just natural language text. It has been shown that some LLMs can read and write computer code, understand and generate images, receive and produce structured data (in any specific format), and so on. This is important, because LLMs may need to communicate with various processes within a cognitive architecture that may use symbolic or subsymbolic formats for internal representation and processing.

Fourth, with LLMs added into a cognitive architecture, more interesting interactions among them and with various other components are possible. For example, when LLMs are incorporated into a dual-process architecture, they can lead to synergistic interaction (Sun, 2024).

Thus, considering all these factors, incorporating LLMs into Clarion seems a solid idea. Below, we examine how LLMs may be incorporated into Clarion.

## 3.2. Inner Working of LLMs

To incorporate LLMs into Clarion, we need to better understand their possible roles in the overall architecture of the human mind. When characterizing LLMs, one may consider implicit mental processes in general and human intuition and instinct in particular (Cosmides & Tooby, 1994; Dreyfus & Dreyfus, 1986; Goldberg, 1983; Sun & Wilson, 2014). Here, intuition refers to understanding or thoughts without explicit reason (Sun & Wilson, 2014), while instinct refers to patterns of responses to environmental stimuli without involving explicit reason (McFarland, 1989). In this regard, one may argue that LLMs correspond roughly to human intuition and instinct (setting aside peripheral implicit processes in perception, motor control, language, and so on; Sun, 2024).

Why should LLMs correspond to intuition and instinct? That is, what characteristics of LLMs make them so? Let us examine intuition (briefly for length considerations; see Sun, 2024 for details). Human intuition is unconscious/implicit (Evans & Frankish, 2009; Kahneman, 2011; Reber, 1989), but it may be viewed as a form of reasoning (Sun, 1994): Reasoning encompasses explicit processes on one hand, and implicit processes (intuition) on the other (with both directed by one's motives, goals, etc.). In fact, intuition is arguably indispensable elements in human thinking — They supplement or even guide explicit reasoning. Empirical work demonstrating the above point can be found in Evans and Frankish (2009), Reber (1989), and so on, while computational arguments includes Sun & Wilson (2014), Helie & Sun (2010), and others.

The processes leading up to intuition are often somehow mysterious and beyond language (Dreyfus & Dreyfus, 1986; Goldberg, 1983; Sun, 1994). However, intuition is (at least to a significant extent) expressible through language. Given a massive amount of linguistic data, it is likely that human intuition about a vast array of subjects may be contained in it. It may conceivably be argued that LLMs trained with such data can gain a large range of intuition, somewhat comparable to humans. That is, a vast number of sentences from a vast number of different sources can capture a large amount of human intuition collectively.

On top of that, the training process with gradual numerical weight adjustments (in Transformer models, as the basis for LLMs) leads to generalization, which in turn may lead to better intuition of an even broader scope. Conceivably, LLMs can embody intuition generalized from a large volume of data.

Furthermore, human intuition results from and relies on diverse (mostly implicit) commonsense knowledge about how the world works, including, for example, naive physics, folk psychology, folk sociology, and so on. This is what a massive amount of text training data from diverse human sources, which inevitably embody such knowledge, provides to LLMs. In addition, it has been shown that a commonsensical mental model of the world may be recovered (inferred) by LLMs from text (Yildirim & Paul, 2024; see also Chang & Bergen, 2023).

It has been noted that human intuition often results from grasping underlying statistical patterns and structures of the world in various respects, in an implicit way, from repeated experience (Hasher & Zacks, 1979; Reber, 1989). Likewise, it has been shown that repeated training of LLMs with a large amount of data facilitates implicit capturing of statistical patterns and structures of the data (e.g., Durt et al., 2023; Mollo & Millière, 2023; Titus, 2023).

Empirical testing of LLMs provides some preliminary indications that LLMs do capture human intuition. For instance, Dasgupta et al. (2022) and Saparov and He (2022) showed that LLMs demonstrated content effects in reasoning similar to humans. Trott et al. (2023) showed that LLMs could capture human intuition exhibited in false-belief experiments. Many other cases were discussed in, for example, Chang & Bergen (2023), Sartori & Orrù (2023), Sun (2024), or Yildirim & Paul (2024).

Some may claim that there is a fundamental difference between intuition and language expressing intuition, or between intuition gained from the "real" world and whatever gained from text. Such claims seem to have divided language and the world in a radical and unjustifiable way. To any individual, the world is made up of sensory-motor, linguistic, and other experiences, which often mirror each other (see, e.g., Pavlik, 2023; Søgaard, 2022). Thus, one may gain intuition from any of these modalities, as well as from their combination. Note that word meanings often emerge from use patterns (Durt et al., 2023; Titus, 2023), and thus it is possible to obtain meaning in LLMs (even from text alone). As argued by Mollo & Millière (2023), a "proximate" function, such as next-token prediction in training LLMs, may give rise to an "ultimate" function, such as grounding or intuition. Yildirim & Paul (2024) similarly discussed how (implicit) world models and knowledge could be inferred/recovered from mere next-token prediction training. See Sun (2024) for discussions of related philosophical issues.

Besides intuition, what about instinct (another type of implicit processes)? Sun and Wilson (2014) discussed how implicit action selection (as guided by one's motives and goals) capture instinct. Beyond hardwired (innate) instincts, there are also acquired (learned) instincts, mostly through repeated experiences. What was argued earlier regarding intuition can be largely applied to instinct. In other words, instinct can develop within LLMs through experiences with a large amount of data.

However, can LLMs capture more than implicit processes? LLMs do have some symbolic processing capabilities, including handling stepwise reasoning, structures, and compositionality to some extent (for early arguments, see, e.g., Pavlick, 2023; Saparov & He, 2022). However, it is worth pointing out that LLMs usually have only limited such capabilities, while human implicit mental processes have been shown to likewise have limited such capabilities (Macchi et al., 2016; Reber, 1989). It seems unnecessary to posit anything more than implicit processes to explain limited symbolic capabilities of LLMs.

Some may object that some LLMs (such as ChatGPT o4 or DeepSeek R1) can be trained to perform very well on reasoning tasks, including complex mathematical reasoning, so they go beyond intuition. In this regard, one should note that human intuition can likewise be trained to perform well on logical or mathematical reasoning tasks. For instance, a well-trained logician can be capable of reasoning logically even without consciously involving explicit step-by-step procedures. A well-trained mathematician may be able to see the mathematical implications of a hypothesis even before any explicit analysis. Likewise, a chess master may see a good move before analyzing a myriad of possibilities (Dreyfus & Dreyfus, 1987). So, a good reasoning performance alone is not indicative of involving more than intuition.

### 3.3. Possibilities of Incorporating LLMs

For incorporating LLMs into Clarion, there are many possible roles for LLMs consistent with the analysis above, ranging from LLMs being at the periphery of Clarion to LLMs being at its core (Romero et al., 2023; Sumers et al., 2023):

- For dealing with language-based communication: In this case, natural language inputs are directed to LLMs (properly fine-tuned), and LLMs process them (implicitly) to provide inputs to Clarion (in its internal representation). Clarion can then process them; based on the outcome, LLMs generate natural language texts as a response.
- For dealing with perception: Multimodal LLMs (properly fine-tuned) can handle implicit (early-stage) perceptual processes. Given a raw input (e.g., a visual scene), LLMs may generate a verbal description (or a form of Clarion internal representation) to be processed by Clarion.

- For dealing with motor processes: LLMs (properly fine-tuned) may generate motor commands, based on higher-level directives from Clarion.
- For capturing memory: In this case, some modules of Clarion may be based on LLMs, which serve as implicit memory of various forms in Clarion (as described earlier; Sun, 2016).
- For capturing all forms of implicit processes: Multiple LLMs (each properly fine-tuned) capture implicit processes ranging from intuition to instinct (as discussed before), aside from implicit processes of perception, motor control, memory, and natural language communication (as mentioned above).

The last approach above evidently confers the most extensive roles on LLMs, as it essentially encompasses all the other possibilities mentioned. This approach is preferrable due to its generality. This approach is also well justified theoretically (see Section 2.2; see also Sun, 2024). Moreover, this approach is readily applicable to any cognitive architecture that allows for dual processes. See Figure 2 for a sketch of the new Clarion with LLMs incorporated (as will be further explicated below).

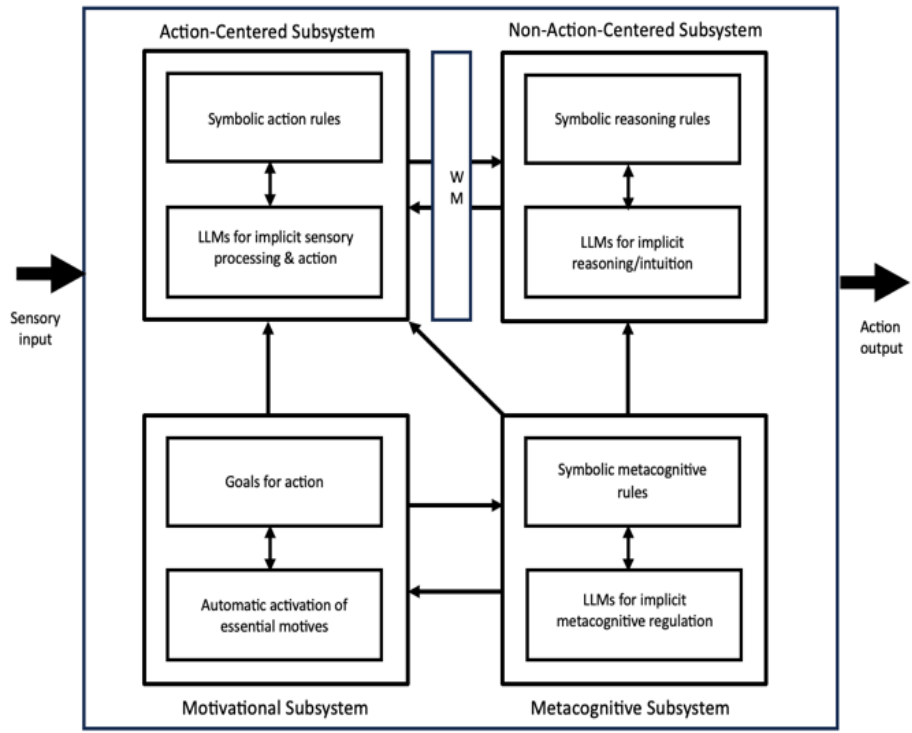

Figure 2. The Clarion architecture incorporating LLMs.

With this last approach, the roles of the two levels are naturally well defined. On the one hand, the roles of LLMs (each fine-tuned, capturing a different implicit process) include, among others:
- Natural language processing (in an implicit way)
- Intuition, including intuitive reasoning, intuitive metacognitive reflection, and so on (possibly directed by explicit processes)
- Implicit action selection (in an instinctual way, possibly directed by explicit processes)
- Implicit memory stores (e.g., semantic, procedural, and episodic memory; Sun, 2016), and implicit mental models of the world constituted from implicit memory (which are especially useful for internal mental simulation of the world)
- Learning from explicit processes (e.g., assimilating explicit knowledge into LLMs); helping explicit processes to learn (e.g., through knowledge from implicit processes)

On the other hand, the roles of explicit processes include, among others:
- Directing natural language processing by LLMs (as needed)
- Directing implicit reasoning by LLMs (as needed)

- Directing implicit action selection by LLMs (as needed; usually in familiar or well-practiced situations)
- Precise, explicit, rule-based and/or logical reasoning (in addition to intuitive reasoning by LLMs)
- Explicit action selection and planning (e.g., in unfamiliar situations) (in addition to instinctual behaviour from LLMs)
- Explicit memory stores (e.g., semantic, procedural, and episodic memory), and explicit mental models of the world resulting from explicit memory
- Explicit metacognitive reflection
- Learning from implicit processes of LLMs (e.g., through extracting their knowledge); helping LLMs to learn (e.g., through providing knowledge that LLMs can assimilate)

For example, in one scenario, the top level does the following: (1) providing inputs to the LLMs to initiate their processing; (2) verifying, rectifying, refining, or explaining the results from the LLMs, using explicit knowledge; (3) or even supplanting the results from the LLMs, using explicit knowledge (when situations call for it). We will examine some of these roles below.

### 3.4. Possibilities of Cross-level Interaction

First, communication between implicit and explicit processes needs to be addressed. As discussed earlier, during processing, they influence each other through dynamic interaction. In addition, top-down and bottom-up learning, as described earlier, involve interaction of the two levels.

In terms of representational form used in the communication between the two levels, there are a number of possibilities. As noted earlier (in 2.3), internal symbolic representations in the original Clarion can be in the form of chunk nodes at the top level, each of which represents a concept and may serve as either a condition or a conclusion in explicit rules. Communication can occur between these chunk nodes at the top level and the corresponding microfeature nodes at the bottom (which "define" each of these chunk nodes; Sun, 2016).

However, with LLMs incorporated, output of LLMs at the bottom level can be:
- Linguistic (natural language) representation
- Embeddings (vectors)

- Chunk nodes as in the original Clarion
- Microfeatures (which may activate chunk nodes, partially or fully)
- Extensions of the preceding possibilities, such as some form of knowledge graphs

Correspondingly, representation used at the top level can be:
- Linguistic representation
- Simple rules with chunk nodes as in the original Clarion
- Extensions such as knowledge graphs

Thus, more flexibility results when LLMs are incorporated into Clarion. Representation involved can be either (1) subsymbolic (i.e., embeddings), (2) symbolic (e.g., linguistic representation), or (3) in the internal Clarion form (from the original Clarion, which involves both symbolic and subsymbolic components; Section 2.3). In this regard, note that LLMs can translate between internal or subsymbolic representation and full linguistic representation (Romero et al., 2023). Such translation allows for the processing of natural language inputs (e.g., queries, instructions, or situational descriptions) into internal forms and the generation of outputs from internal forms.

However, for the purpose of communication between the two levels, using internal Clarion representation (as from the original Clarion) is possible but, when LLMs are involved, too limiting. Using embeddings from LLMs is complex and would require a much longer, more specialized treatment, thus omitted here. The third approach, using linguistic representation, is the most natural for LLMs, which also works for the top level (i.e., it can adopt such representation too). Therefore, we will explore this approach here.

Next, when LLMs are incorporated, interactive dynamics of the two levels elucidated in the original Clarion may be mapped to those between LLMs and explicit processes. That is, the Clarion way of interaction may be leveraged for structuring the interaction between LLMs and symbolic models. Specifically, with linguistic representation being used for cross-level communication, top-down and bottom-up activation originally specified in Clarion become (1) prompts in a linguistic form from the top level to the LLMs, and (2) outputs from the LLMs in a linguistic form, which are then incorporated into the top level. Using prompts means prompt engineering and other techniques associated with LLMs become relevant. Furthermore, bottom-up and top-down learning of the original Clarion can also be mapped to the new Clarion (see Appendix).

Note that, within the new Clarion, the interaction through prompting, in a way, resembles an inner dialogue within oneself. In this case, the dialogue mainly consists of repeated queries (from explicit processes at the top level) to one's inner, deeper self (implicit processes at the bottom level). Through this dialogue, one's thoughts form, develop, and evolve. As pointed out by Vygotsky (1962), as one internalizes inter-personal communicative utterances, inter-personal dialogue becomes intra-personal inner dialogue. In this case, one's explicit mind is engaged in a conversation with one's implicit mind. Linguistic representation is perfectly suited for this conversation, as the explicit mind has a more active and more dominant role.

## 3.5. Incorporating LLMs into Clarion

### 3.5.1. An Example with Linguistic Cross-level Interaction

One version of the new Clarion, relying on linguistic representation for cross-level interaction, is illustrated with the examples below. (For a more complete specification of the model, see Appendix.)

Imagine a linguistic input is received by a Clarion agent from an external source (e.g., a person), which is a request to perform a task, such as "could you bring me a knife?". Assume the agent is currently at an unfamiliar house, regarding which only some limited information is available. At this point, the top level of the ACS of the agent would search its explicit knowledge to recommend an action, which leads to failure, as there is no relevant knowledge available. At the same time, the top level of the ACS would also prompt the LLM (at the bottom level) to recommend an action (see Appendix). Due to a similar lack of knowledge at the bottom level, it comes back with a general suggestion: Let's think about this (i.e., "in-step reasoning using NACS"; see Appendix; Yao et al., 2023). A directive is issued to the NACS, along with the goal of the task and the current situation (i.e., the partial action trajectory thus far, which consists only of the initial state in this case), using the working memory to convey all the information. Thus, the NACS is invoked, and its explicit and implicit knowledge is utilized. Using explicit knowledge at the top level fails (due to lack of relevant knowledge). At the same time, the top level prompts the bottom level (consisting of LLMs) for in-step reasoning with all pertinent information (i.e., the partial action trajectory so far). Due to the lack of specific knowledge of the house (despite its vast intuitive knowledge about the world), the bottom level, instead of providing specific information, responds with a plausible suggestion "try the kitchen of the house" (because a knife is likely found in a kitchen). This suggestion is sent back to the ACS (via the working memory), and the ACS gets back to action selection. The ACS then recommends going to the kitchen as a first step. Then, after some intermediate steps, once arrived in the kitchen, the agent repeats essentially the steps above and narrows down where in the kitchen a knife might be, and so on, until a knife is found. After some more steps, once the agent finally accomplishes the goal of bringing the knife to the person who requested a knife, it will receive an internal reward, determined by the MCS based on overall satisfaction of activated drives in the MS (e.g., motives such as *achievement*; their satisfaction results from achieving the goal; Sun et al., 2022).[1]

Then learning occurs. At the bottom level of the ACS, reinforcement learning based on the said reward takes place in the LLMs (so that the agent can perform better next time, same as in the original Clarion). Further, bottom-up learning extracts symbolic rules from the bottom level of the ACS and adds them to the top level of the ACS (essentially the same as in the original Clarion, so that they can be accessible, e.g., to provide explanations when needed; see Appendix for details). Other types of learning are also possible, such as learning explicit knowledge in a linguistic form at the top level from external sources (when available; such as instructions from a person), and then top-down learning (assimilation of such knowledge into the bottom level, e.g., through reinforcement learning guided by the top-level knowledge; Appendix). Similar learning also occurs in the NACS. [2]

If a task can be repeated, then the agent may, over time, improve its performance iteratively. For instance, on the first trial, the agent may fail to achieve the task goal. But, after some iterations, it becomes capable of achieving the goal but takes too many steps to do so (e.g., beyond a given limit). After more iterations, the agent accomplishes the task within the given constraints. This process is carried out by an iterative loop coordinated by the MCS: If an action trajectory generated by the ACS for a task is not satisfactory, then "end-of-trajectory reflection" is carried out by the MCS (see Appendix; Shinn et al., 2023). The bottom level of the MCS (an LLM) provides a reflection on what the agent did right or wrong and what the agent should try next. The result of the reflection is passed to the ACS (via the working memory). Then a new iteration ensues, which generates a new, better action trajectory, and so on. In the meantime, reinforcement learning also occurs in the LLMs at the bottom level of the ACS, based on the reward from the MCS (the overall drive satisfaction). Other types of learning can also occur.

---

[1] If an input is visual, it must be converted to linguistic descriptions for the top level of the ACS (converted by an LLM, as a rough stand-in for a visual system), but it might be directly used by the bottom level of the ACS (i.e., LLMs) as part of its input (along with a linguistic prompt).

[2] Note that, as indicated in the description above, the NACS is usually controlled by the ACS, and in turn the ACS is often controlled by the MCS (see Sun, 2016 for details).

Beyond performing a physical task, the Clarion agent can also perform a mental task such as problem solving. For instance, the agent is given a problem to be solved: "How do you plant four trees in such a way that they are at the same distance from each other?" In this case, the ACS selects an action: "general reasoning using NACS" (as the problem requires thinking), determined by the top or the bottom level of the ACS (see Appendix). The problem description and the goal are passed to the NACS via the working memory. Once invoked, the top level of the NACS, besides using its explicit knowledge to reason, also triggers the bottom level of the NACS through a prompt directing it to reason using applicable reasoning methods, such as reasoning by divide-and-conquer, reasoning by analogy, and so on (Xie et al., 2023). The results are integrated (by an LLM, with an integration prompt) and an integrated solution is proposed. If the proposed solution is not considered good enough, then another reasoning cycle may be initiated, with the previous outcome added to the input; if it is considered acceptable, then it is reported back to the person requested the solution (as coordinated by the ACS and MCS). Learning occurs as before.

### 3.5.2. Other Essential Elements and Their Interactions

Beside those interactions as illustrated by the examples above, there are other important interactions that are also crucial to the working of Clarion. For instance, there are a variety of memory stores in Clarion, as discussed earlier, and they interact with each other and with other components to accomplish various functions.

#### 3.5.2.1. Memory

LLMs, given their vast capacities for implicit knowledge, serve well as implicit semantic memory and implicit procedural memory within Clarion, at the bottom levels of the NACS and the ACS, respectively (see Appendix), while the top levels of the NACS and the ACS serve as their explicit counterparts (Sun, 2024; cf. Schacter, 1987). Therefore, LLMs serving as these forms of implicit memory interact with each other and with their explicit counterparts.

However, episodic memory (Tulving, 1983) and working memory (Baddeley, 1986) are also important psychologically, although usually absent inside of LLMs. Clarion provides these memory stores (cf. Sumers et al., 2023). In Clarion, explicit episodic memory consists of one's experiences, in the form of a list of n-tuples: *state, action, reinforcement*, and so on (along with time stamps). A retrieval function takes a partial description of state, action, and so on as input and returns a subset of the memory that matches the input (e.g., to use as the basis for reasoning or learning). in addition, a variety of techniques were developed for condensing, summarizing, and integrating memory with LLMs (e.g., Park et al, 2023), which in turn lead to abstract episodic memory (in a linguistic form, thus also residing within the explicit episodic memory; Appendix).

In addition to explicit episodic memory, implicit episodic memory in Clarion is formed based on its explicit counterpart: Neural networks are used for storing "compacted" episodes (in an implicit way), which can predict, for example, action based on current state, next state based on current state and action, and so on, for forward planning and simulation. Thus, episodic memory is also dual-representational (with each part serving a different function).

Episodic memory may be utilized for learning within the ACS or the NACS. For instance, episodes retrieved may be used for fine-tuning LLMs. Episodes retrieved can also be used for in-context and instance-based learning (more later).

Furthermore, working memory in Clarion stores information on a temporary basis, facilitating the processing of the stored information (especially useful for the ACS), or for transferring information (especially between the ACS and the NACS; Sun, 2016). Such working memory applies to the new Clarion, to transfer information between subsystems (as in the original Clarion) and to remedy the lack of memory within LLMs.

### 3.5.2.2. Motivation and Metacognition

Other subsystems in Clarion (besides the ACS and the NACS) and their interactions with the ACS and the NACS should be further sketched here. In Clarion, as discussed earlier, the MS provides essential, intrinsic motives (i.e., drives) as the basis for self-direction and self-regulation of behavior (carried out by the MCS). Work on Clarion shows that drives are essential to accounting for a broad range of human behavior, ranging from work performance to moral judgment (e.g., Bretz & Sun, 2018; Sun et al., 2022). That is, the MS has been shown fundamental to the human mind.[3] In the new Clarion, drive activations can be accomplished through a fine-tuned LLM (which is more realistic than the original implementation; cf. Sun, 2016).

Based on drives, goals, and other factors, the MCS monitors and regulates cognition and behavior (Reder, 1996), through interacting with other subsystems. Metacognition and regulation can occur implicitly through LLMs (e.g., Park et al., 2023). Meanwhile, explicit metacognition and regulation can also occur at the top level of the MCS. By capturing metacognition, Clarion can explain a larger range of human behavior.[4] Together, motivation and metacognition, and consequent self-direction and self-regulation, make Clarion a more complete model of the human mind.

### 3.5.2.3. Learning

For learning, interaction among components within Clarion is also important. For example, as explained earlier, bottom-up learning (extracting rules from the bottom level and adds them to the top level) involves interactions between the two levels (within the ACS or the NACS), which is applicable to this new Clarion. An additional opportunity offered by incorporating LLMs is that LLMs themselves may be prompted to generate useful symbolic rules (and other symbolic content) for symbolic processes at the top level (see Appendix). This can be done right after pretraining of LLMs or during continuous learning (thus extracted content reflects continuous learning). The generated symbolic content can add to explicit semantic or procedural memory (at the top level of the NACS or the ACS, respectively), supplementing explicit knowledge obtained from externally provided instructions, external tools and sources, bottom-up learning, and so on.

In addition, other forms of learning can also occur through various interactions. Learning actions from mistakes can be accomplished through interaction between the episodic memory and the ACS, by retrieving cases of past action mistakes from the episodic memory to be used as part of inputs to the ACS (e.g., Xie et al., 2023). Instance-based learning in general can occur (in an in-context way) based on episodes from the episodic memory (as part of inputs to the ACS,

---

[3] By being situated within Clarion, LLMs become also equipped with human-like motivation, which, for LLMs, can enhance value alignment and mutual intelligibility with humans.

[4] By being situated within Clarion, LLMs become also equipped with metacognitive capabilities.

NACS, or MCS). These episodes can also be used, for example, for fine-tuning LLMs. Curriculum learning (e.g., Elman, 1993) can be accomplished in the ACS or NACS under the direction of the MCS. See Appendix for more details.

## 4. Final Remarks

In order to deal with both real-world complexity and psychological realism, a pressing challenge facing cognitive architecture research is to develop cognitive architectures with LLMs (and/or Transformers) playing significant roles. LLMs, due to their powerful computational capabilities, may serve as the key to moving forward such research. In particular, LLMs can be incorporated into Clarion to better capture human intuition, instinct, and other implicit processes, as well as to enable natural language communication, to further enhance Clarion. Integrating Clarion and LLMs may, in some way, also help develop AI systems that are more capable and more human-like (Sun, 2024). In any case, computational cognitive architectures augmented by LLMs have capabilities beyond any previous cognitive architectures before the advent of LLMs. As a result, LLMs-enhanced cognitive architectures may be used for dealing with real-world situations, as opposed to being limited to simulating small laboratory tasks as before.

Other advantages of the new Clarion include: (1) integrating different functionalities (action, reasoning, memory, motivation, metacognition, etc.) into one unified framework with fine-tuned and more capable LLMs playing key roles in each; (2) employing a variety of representational/computational methods (symbolic, neural, and LLMs) for action selection, for reasoning, and so on, leading to better capabilities; (3) including a motivational subsystem that determines underlying motivation for behavior and a more capable metacognitive subsystem that regulates other processes on that basis.

Furthermore, advantages of the new Clarion, as compared to an LLM by itself, includes these three advantages above, plus that of structuring all of them in a cognitively/psychologically plausible and validated way (leveraging prior validation of the original Clarion; Sun, 2016), including foundational psychological structures such as implicit vs. explicit processes, ACS vs. NACS, MS vs. MCS, and so on.

In turn, benefits for LLMs themselves from this combination include: (1) integrating separately tuned LLMs for different functions as modules, with each specialized and thus more effective; (2) unpacking long sequences with intertwined action-reasoning cycles in LLMs (cf. Yao et al., 2023), making the action-reasoning combination easier to handle (alleviating the difficulty of learning sequencing); (3) adding symbolic processes in addition to LLMs; (4) adding intrinsic motivation and metacognition in a psychologically plausible and validated way (Sun et al., 2022).

Yet another point relates to the perennial issue of nature vs. nurture (or nativism vs. empiricism): Some believe that (almost) everything cognitive is learned from experience, while others believe that (most) mental contents have innate origins. However, the most likely scenario is somewhere in between (Buckner, 2024): Some innate structures, mechanisms, and tendencies are likely built-in a priori; learning occurs on that basis. That is where Clarion stands, avoiding radical positions on either end (so as to alleviate, practically speaking, incidental correlation, illusory role, etc. often associated with radical empiricist approaches).

Together, the new Clarion, with LLMs incorporated, serves as a case study of integrating LLMs into cognitive architectures. This case clearly demonstrates that the integration can be natural, seamless, and principled, and the result might be beneficial to both cognitive architectures and LLMs. The work advocates an eclectic, multifarious approach towards better models for cognitive science.

# Appendix

Below is the brief (partial) specification of one version of Clarion that incorporates LLMs through linguistic cross-level interaction. See Table 1 for the notations used in the specification. See the references cited below for further details.

| ACS = action-centered subsystem | NACS = non-action-centered subsystem | MS = motivational subsystem | MCS = metacognitive subsystem |
|---|---|---|---|
| TL = top level | BL = bottom level | TD = top-down | BU = bottom-up |
| WM = working memory | EM = episodic memory | SM = semantic memory | PM = procedural memory |
| RL = reinforcement learning | BLA = base-level activation | | |

Table 1. Notations and conventions used in the new Clarion specification.

**ACS:**

- For each task with a goal, iterate the generation of action trajectories until success or giving up (e.g., after too many iterations). For each of these iterations, repeat the generation of action steps until success or giving up (after too many steps). These two loops are directed by the MCS (specified later). For the ACS, at each action step:
- TL:
  - Select an external action, based on explicit procedural knowledge (i.e., action rules) and relevant input information (specified below). Or, select an executive action "in-step reasoning by the NACS on the current ongoing action trajectory" (*in-step reasoning by NACS*); or, when at the end of an action trajectory, select an executive action "metacognitive reflection by the MCS on the just completed action trajectory" (*end-of-trajectory reflection*); or, select an executive action "general reasoning by the NACS [using method X] [based on information Y] [for objective Z]" (*general reasoning by NACS*)
  - Alternatively, select an action that is a directive (i.e., prompt) to the ACS BL (with relevant input information specified below), which selects action (as specified below)
  - Note:
    - An action can be an external action or an internal (e.g., executive) action.
    - Input information used for action selection by the ACS includes: the current state, which includes perceptual inputs (often described in a linguistic form), the current goal, and the WM. It may optionally include: information from the NACS (through the WM; such as results of in-step reasoning), metacognitive reflection from the MCS (through the WM) regarding immediately preceding action trajectories for the current task (and/or such trajectories themselves and their outcomes/feedback, from the EM), information (through the WM) selected from the EM in general (e.g., past cases of success/failure) and from the SM (e.g., information about an event/person/object)

- o Assemble all such input information first before attempting to choose an action (Xie et al., 2023), except purely reactive responses in which case only perceptual inputs matter. The TL may consider only a limited subset of the information above, while the BL may consider a broader range
- ◆ BL (LLM):
  - Select an action (through its implicit processes), after receiving a prompt from the ACS TL
  - Pre-trained/fine-tuned with implicit procedural knowledge embedded in weights
- ◆ TL/BL Integration:
  - Stochastically choose one level or the other (Sun, 2016); that is, the choice is between top-level action selection (if there is knowledge there to do so) and a directive/prompt to the ACS BL (consequently the bottom-level action selection)
- ◆ Learning:
  - TL: acquiring action knowledge (in a linguistic form) from external sources, translated into action rules; BU learning
  - BL: pre-training/fine-tuning of LLMs; RL on the fly; RL with off-line replay of the EM; in-context learning (e.g., in-step reasoning by the NACS, end-of-trajectory reflection by the MCS, episodes from the EM); TD learning
  - Note:
    - o At each step, an external action from the ACS may change the world, either through the action executed in the external world or in a simulated world (in a mental simulation), or an internal action that change the internal information
    - o Reinforcement might be generated on that basis by the MCS reinforcement module, at every step, at some steps, or only at the end of a trajectory; it is determined through evaluating (by the MCS) the overall satisfaction of activated drives within the MS (based on states

of the external or simulated world; Sun et al., 2022), or from an evaluation LLM (fine-tuned as the reinforcement module)
- TD learning: assimilation of TL action knowledge through RL-based fine-tuning of the BL (Sun, 2016), or in-context learning via prompts (Romero et al., 2023)
- BU learning: (a) first rule extraction from the BL; (b) then rule refinement. This learning possibly considers only a subset of information (e.g., no end-of-trajectory reflection, or no past episodes)
    - An extraction algorithm may be as described in Sun et al. (2001); or through prompting LLMs to extract (e.g., Yang et al., 2024)
    - Refinement (simplified from Sun, 2016): (a) generalization through testing each current condition element of a rule to see if it can be removed (using statistical measures; Sun, 2016), when the rule is successful; (b) specialization through going back to a previous version (that led to the generalization to the current version of the rule), when the rule is not successful; (c) deletion, if no such previous version to go back to, when the rule is not successful

**NACS:**
- ◆ TL:
    - Store explicit declarative knowledge in a linguistic form. Reason based on knowledge at the TL and inputs provided in the WM (e.g., by the ACS). Inference at the TL can be carried out by practically any inference engine
        - For example, a production rule system may be used whereby knowledge at the TL is first translated into production rules and reasoning is then carried out. The same goes for a mental logic system (Braine & O'Brien, 1998), or a theorem prover (Bringsjord, 2008), or whatever
    - At the same time, send to the BL one of the followings (as directed by the ACS/MCS):

- o A prompt to BL for "in-step reasoning" (if directed to do so by the ACS): Direct the NACS BL to (implicitly) reason over the current, ongoing action trajectory (Yao et al., 2023), which includes the overall goal, initial and subsequent states, actions in these states, and so on (from the explicit EM), plus (optionally) proposed actions in the current state. Store the outcome (in a linguistic form) in the WM so that ACS can access. Store also in the explicit EM (as part of the action trajectory)
- o A prompt to BL for "general reasoning" (if directed to do so by the ACS): Direct the NACS BL to reason (implicitly) from inputs provided in the WM, using one or more of the available reasoning methods (as determined by the ACS or MCS, such as divide and conquer, analogy, etc.; Xie et al., 2023). If multiple methods are used, integrate their results (e.g., by using an integration LLM). Store the outcome (in a linguistic form) in the WM so that the ACS can access

◆ BL (LLM):
- Carry out reasoning as directed by the NACS TL (based on the guidance from the ACS or MCS)
- Pre-trained/fine-tuned for such reasoning

◆ TL/BL Integration:
- Union of results from the two levels (which work in parallel and simultaneously)
- Then, optionally, integration of the results (e.g., by using an LLM for a more holistic integration, or an inference engine for a more analytical integration verifying, enhancing, or explaining results)

◆ Learning:
- TL: acquiring knowledge from external sources (in a linguistic form); BU learning
- BL: pre-training/fine-tuning of LLMs; TD learning; learning from past episodes (supervised fine-tuning)

- TD learning: assimilation of knowledge from the TL (through supervised fine-tuning; Sun 2016), or in-context learning via prompts
- BU learning: extraction of knowledge from the BL, and then refinement of the extracted knowledge
  - An extraction algorithm for the NACS may be as described in Killian and Sun (2025) or Sun (2016), or through prompting an LLM (Yang et al., 2024)

**MS**

- ◆ BL:
  - Consisting of a set of primary drives (as specified in Sun, 2016; Sun et al., 2022), which are activated based on perceptual inputs and internal inclinations (e.g., through a fine-tuned LLM)
- ◆ TL:
  - Consisting of goals (set by the MCS or ACS), with one of them being the active goal (e.g., selected based on BLAs; Sun, 2016)

**MCS control module**:

- ◆ TL:
  - Store explicit metacognitive control knowledge and perform the following functions
  - For iterative generation of action trajectories in the ACS: For each task (with its goal), direct the ACS to repeat the generation of action trajectories (each leading to a trajectory — a sequence of actions for accomplishing a task), until success (i.e., finding an action trajectory achieving the task goal, within specified constraints) or giving up (e.g. after too many iterations without success). For each of these iterations, repeat the selection of action steps until success (achieving the task goal) or giving up (after too many steps without success). For tasks where iteration is not possible in the world, iterative generation of action trajectories in a mentally simulated world may be attempted, or no iteration at all

- For end-of-trajectory reflection: At the end of an action trajectory, direct its BL to metacognitively reflect on the most recently completed action trajectory (or several recently completed action trajectories) from the EM and generate insight (in a linguistic form) regarding correct/incorrect steps/elements therein (as in Shinn et al., 2023). Note that a completed action trajectory includes: the overall goal, initial and subsequent states, actions in these states, and so on, along with rewards (final and intermediate, if any). Store the result of reflection (in a linguistic form) in the WM, to be used by the ACS to generate further action trajectories. Store also in the explicit EM (as part of the action trajectory reflected on)
- For NACS: When the NACS is invoked for reasoning, if its operational details are not specified by the ACS, direct the NACS to use one or more of the reasoning methods (as mentioned earlier) and, if more than one method are used, to integrate their results (e.g., by using an integration LLM). Store it (in a linguistic form) in the WM
- For NACS: For memory retrieval within the NACS, assess the following measures for each relevant piece of information (from the EM or SM): (a) importance (e.g., using an evaluation LLM), (b) relevance (e.g., using embedding similarity), (c) recency (using BLA), and so on (Park et al., 2023), with respect to events/persons/objects involved in the current state. Select information based on the assessment and store in the WM
- For NACS: Optionally, after gathering all pertinent information from the EM and/or SM (e.g., by the methods above), direct the module BL to reflect on it and generate insight about an event/person/object (Park et al., 2023). Store it in the WM. Store it also in the explicit SM (in a linguistic form)
- For NACS: Optionally, after the completion of several tasks by the ACS, direct the module BL to summarize experiences with these tasks (at an abstract level), based on information in the EM (e.g., considering only the final action trajectory for each of these recent tasks) (Park et al., 2023).

   Store (in a linguistic form) in the abstract explicit EM (with or without links to actual experiences in the EM)
- ◆ BL (LLM):
  - Pre-trained/fine-tuned for metacognitive reflections
  - Optionally store also implicit metacognitive control knowledge, acquired through TD learning
- ◆ TL/BL Integration:
  - Stochastically choose one level or the other as in the ACS

**MCS reinforcement module**
- ◆ Consisting of a neural network or a fine-tuned LLM carrying out this function (estimating overall drive satisfaction)

**MCS goal setting module**
- ◆ Set goals based on drive activations and relevance parameters (Sun et al., 2022; e.g., through a neural network or a fine-tuned LLM)

**WM**
- ◆ Consisting of a limited number of slots each of which can hold information in a linguistic form, which can come from the NACS or ACS, for either to access
- ◆ Actions on the WM include storing or removing information (by partial matching of content), initiated by the ACS or MCS (Sun, 2016)
- ◆ Relatively rapid forgetting of stored information based on BLA (Sun, 2016)

**SM**:
- ◆ Located within the NACS, as described earlier regarding the NACS (as it is the main part of the NACS)
- ◆ Explicit SM:
  - Located at the TL of the NACS
  - Involving BLA for capturing recency, priming, forgetting (with very slow forgetting rates), and so on (Sun, 2016)

- Content types: (a) basic semantic information, (b) summarized semantic information (resulting from reflections). Different forgetting rates (based on BLA) are involved in these types, from high to low, in that order
 ◆ Implicit SM:
   - Located at the BL of the NACS
   - Same types of content (but intermeshed in weights, in part from pre-training/fine-tuning, in part from TD learning described before regarding the NACS, and so on)

**EM**:
 ◆ Also located within the NACS, separate from what was described above regarding the SM (the main part of the NACS)
 ◆ Explicit EM:
   - Located at the TL of the NACS, separate from the SM
   - Involving BLA for recency and priming, as well as for forgetting (with slower forgetting rates than the WM)
   - Content types: (a) pre-final action trajectories for a task (including reasoning/reflections), (b) the final action trajectory for a task (including reasoning/reflections), (c) abstract (summary of) episodes (in a linguistic form, located in the abstract EM). Different forgetting rates (based on BLA) are involved in these types, from high to low, in that order
   - Before a task is completed, keep all its generated action trajectories (so as, e.g., to be able to reflect on them), each in a sequence of state-action-feedback quadruples (along with reasoning and reflections); after a task is completed, these trajectories will have a higher forgetting rate except the final trajectory
 ◆ Implicit EM:
   - Located at the BL of the NACS, separate from the SM (the main part of the NACS)
   - Consisting of neural networks for predictions of the action (given the state), the state (given the previous state and the action), and so on, learned from the explicit EM (through supervised fine-tuning). This is useful for mental simulation of the world or for off-line learning of the ACS BL